\documentclass[11pt]{article}

\usepackage[preprint]{acl}

\usepackage{times}
\usepackage{latexsym}

\usepackage[T1]{fontenc}

\usepackage[utf8]{inputenc}

\usepackage{microtype}

\usepackage{inconsolata}

\usepackage{graphicx}

\usepackage{listings}

\title{Evaluating Prompting and Execution-Based Methods for Deterministic Computation in LLMs}

\author{
Hongkun Yu \\
Virginia Tech \\
Blacksburg, VA, USA \\
\texttt{yuhongkun@vt.edu} \\
\href{https://orcid.org/0009-0009-8891-7879}{ORCID: 0009-0009-8891-7879}
}

\begin{document}
\maketitle

\begin{abstract}
Large Language Models (LLMs) have demonstrated strong capabilities in natural language understanding and reasoning. However, their ability to perform \emph{exact, deterministic computation} remains unclear. In this work, we systematically evaluate multiple prompting strategies—including Chain-of-Thought (CoT), Least-to-Most decomposition, Program-of-Thought (PoT), and Self-Consistency (SC)—on tasks requiring precise and error-free outputs, such as binary counting, longest substring detection, and arithmetic evaluation.

To support this study, we introduce a synthetic dataset with diverse natural language instructions, enabling controlled evaluation of exact computation across multiple task types.

Our results show that standard prompting methods achieve only moderate accuracy on sequence-based tasks, while CoT provides limited improvement and Least-to-Most suffers from error accumulation. In contrast, PoT achieves perfect accuracy by generating executable code and delegating computation to an external interpreter. Self-Consistency further improves robustness through majority voting but incurs substantial computational overhead.

We additionally train a small domain-specific model (CodeT5-small) to generate executable programs, which achieves perfect accuracy on the synthetic test set across all tasks with minimal training cost.

Overall, our findings suggest that LLMs may simulate reasoning patterns rather than reliably perform exact symbolic computation. For deterministic tasks, combining LLMs with external tools or using specialized models provides a more reliable and efficient solution.
\end{abstract}

\section{Introduction}

Large Language Models (LLMs) have demonstrated remarkable performance across a wide range of natural language processing tasks, including question answering, code generation, and reasoning benchmarks. Despite these advances, their ability to perform \emph{exact, deterministic computation} remains limited. Tasks such as counting, symbolic manipulation, and arithmetic evaluation require precise and error-free execution, which fundamentally differs from the probabilistic nature of language modeling.

Recent work has explored prompting techniques to improve reasoning capabilities in LLMs. Methods such as Chain-of-Thought (CoT), Least-to-Most decomposition, Program-of-Thought (PoT), and Self-Consistency (SC) have shown success on reasoning benchmarks. However, these approaches are primarily evaluated on tasks where approximate reasoning is sufficient, and their effectiveness on \emph{exact computation problems} remains unclear.

In this work, we systematically study the performance of LLMs on deterministic computation tasks. We focus on three representative problems: (1) counting the number of 0s and 1s in binary sequences, (2) finding the longest consecutive run of a given character, and (3) evaluating arithmetic expressions. These tasks require exact outputs and provide a controlled setting for analyzing the limitations of LLM-based reasoning.

We evaluate multiple prompting strategies, including CoT, Least-to-Most, PoT, and SC, and compare them with a small domain-specific model (CodeT5-small) trained to generate executable programs. Our experiments reveal that prompting alone is insufficient for reliable deterministic computation, while program-based approaches and lightweight supervised models achieve significantly better performance.

Our contributions are threefold:
\begin{itemize}
\item We construct a diverse synthetic dataset for evaluating exact computation across multiple task types.
\item We provide a systematic comparison of prompting strategies under deterministic settings.
\item We demonstrate that program-based approaches and small specialized models can reliably outperform large LLMs on exact computation tasks.
\end{itemize}

We release our dataset and evaluation code to facilitate further research in this area.

\section{Related Work}

Large Language Models (LLMs) have demonstrated strong few-shot and zero-shot capabilities across a wide range of tasks since the introduction of GPT-3 \cite{brown2020gpt3}. However, improving their reasoning ability remains an active area of research.

Chain-of-Thought (CoT) prompting \cite{wei2022chain} encourages models to generate intermediate reasoning steps, improving performance on multi-step reasoning tasks. Self-Consistency \cite{wang2022self} further enhances CoT by sampling multiple reasoning paths and selecting the most consistent answer via majority voting.

Least-to-Most prompting \cite{zhou2022least} decomposes complex problems into simpler subproblems, solving them sequentially to improve reasoning accuracy. Program-of-Thought (PoT) prompting \cite{chen2023program} extends this idea by generating executable code instead of natural language reasoning, allowing models to leverage external computation tools for precise results.

Recent work has also explored tool-augmented language models, where models learn to interact with external tools to improve performance on tasks requiring precise computation \cite{schick2023toolformer}. These approaches highlight the importance of integrating symbolic execution with language modeling.

In parallel, program synthesis and code generation models such as CodeT5 \cite{wang2021codet5} have shown strong performance in generating executable programs from natural language descriptions. These models are particularly well-suited for deterministic tasks where correctness can be verified through execution.

Despite these advances, most prior work focuses on general reasoning benchmarks rather than exact symbolic or deterministic computation. In this work, we systematically evaluate these prompting strategies and compare them with a fine-tuned code generation model under exact computation settings.

\section{Dataset}

The motivation for our dataset comes from exact sequence analysis problems, such as counting bit frequencies in long binary sequences (e.g., in cryptographic contexts). These tasks require exact computation rather than approximate reasoning. Although these tasks appear simple, they require exact deterministic computation rather than approximate language understanding. We therefore extend this idea into a broader benchmark for evaluating whether LLMs can reliably solve symbolic and arithmetic computation tasks.

We construct a synthetic dataset with three task types: binary counting, longest substring detection, and arithmetic computation. Each task is generated automatically with ground-truth answers and executable target programs. For each task, we generate 1000 training examples, 500 validation examples, and 200 test examples.

\subsection{Task Types}

\paragraph{Binary Counting.}
Each sample contains a binary sequence of length 80--120. The goal is to count the number of 0s and 1s separately. We also include special cases where the entire sequence consists only of 0s or only of 1s, making the dataset test both normal and edge-case behavior. The target answer format is:
\[
\texttt{0:<count\_0> | 1:<count\_1>}
\]

\paragraph{Longest Substring.}
Each sample contains a sequence generated from lowercase letters, digits, and special symbols. A target character is randomly selected, and a repeated block of this character is inserted at a random position. The model must return the length of the longest consecutive substring consisting only of the target character. This task tests sequential state tracking rather than simple counting.

\paragraph{Arithmetic Computation.}
Each sample contains an arithmetic expression with 20--40 numbers and operators from \texttt{+}, \texttt{-}, \texttt{*}, and \texttt{/}. Expressions are generated so that the final result is an integer. This avoids ambiguity from floating-point outputs and allows exact-match evaluation.

\subsection{Generalized Instructions}

To avoid testing only memorized prompt templates, each task uses multiple natural language instructions. For example, binary counting instructions include ``Return the separate counts of 0s and 1s'' and ``Determine the frequency of 0 and the frequency of 1''. Longest substring instructions explicitly ask for the length of the longest consecutive block, such as ``Return the integer length of the longest repeated \{target\_character\} segment''. Arithmetic instructions include variants such as ``Evaluate the arithmetic expression'' and ``Use operator precedence to calculate the expression''.

This design makes the benchmark closer to realistic user queries, where the same underlying task may be expressed in different ways.

\subsection{Data Format}

Each example is stored in JSONL format. All tasks share a unified structure containing the task type, instruction, input, output, and target code. The \texttt{task\_type} field allows both prompting methods and fine-tuned models to identify which computation pattern is required.

A binary counting example is shown below:
\begin{verbatim}
{
  "task_type": "binary_count",
  "instruction": "Counting 0s and 1s...",
  "input": "00100000100000001011...",
  "length": 120,
  "output": {
    "count_0": 65,
    "count_1": 55,
    "answer": "0:65 | 1:55"
  },
  "target_code": "seq = input_data..."
}
\end{verbatim}

For the longest substring task, the example additionally includes a \texttt{target\_char} field. For arithmetic computation, the input is an expression string and the target code evaluates it directly.

\subsection{Dataset Statistics}

The mixed dataset contains 1000 training examples, 500 validation examples, and 200 test examples for each task. Token length analysis shows that the longest substring task produces the longest inputs and target programs. Based on this analysis, we set \texttt{MAX\_INPUT\_LEN = 160} and \texttt{MAX\_TARGET\_LEN = 83}, which covers 100\% of the mixed dataset without truncation. The dataset and code are publicly available at: \href{https://github.com/bigbird231/llm-exact-computation-dataset}{GitHub repository: llm-exact-computation-dataset}. All prompts, datasets, and evaluation scripts are released for reproducibility.

\begin{figure}[h]
\centering
\includegraphics[width=1\linewidth]{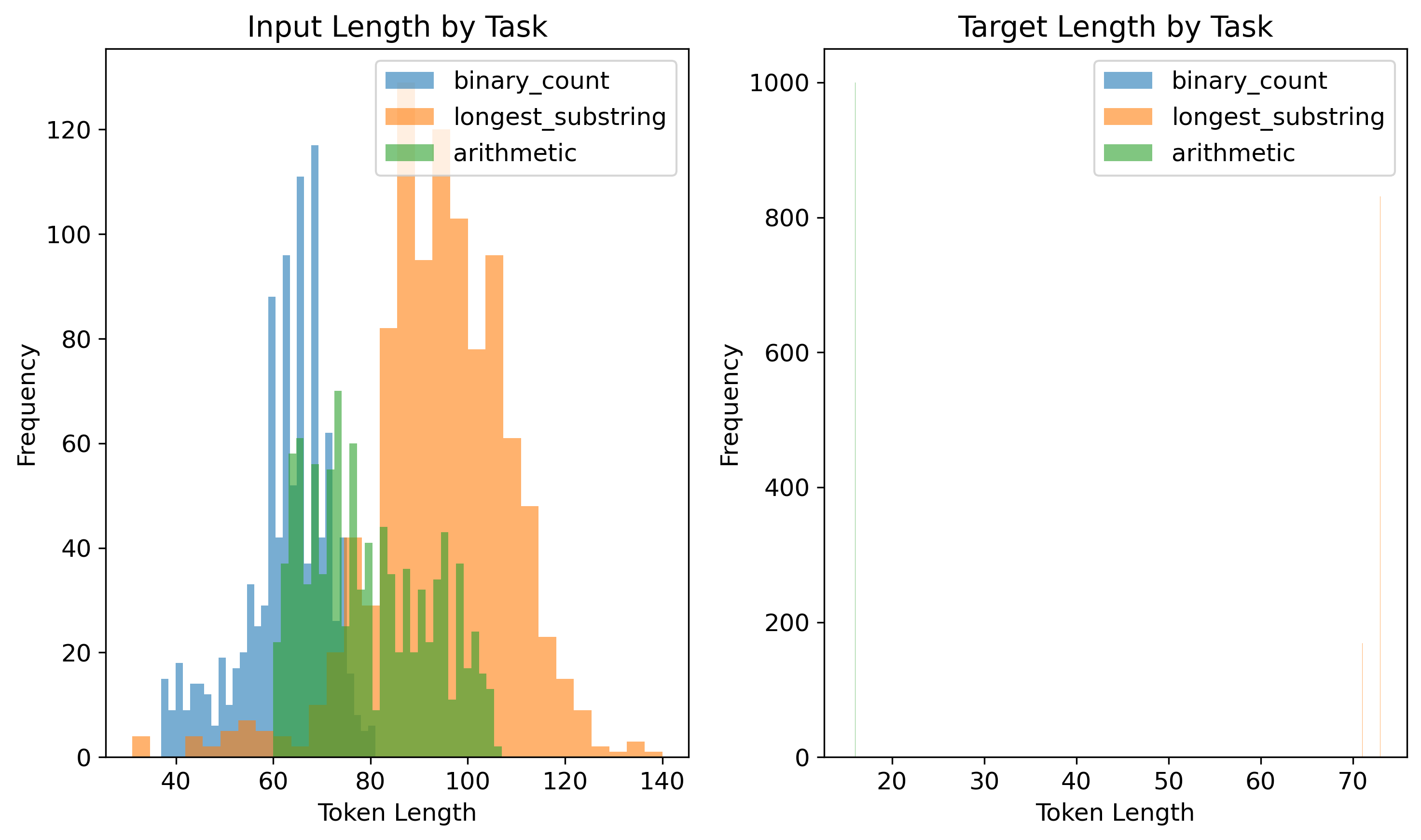}
\caption{Token length distributions of the mixed dataset. The longest substring task produces longer inputs and target programs compared to binary counting and arithmetic tasks, motivating the choice of larger maximum sequence lengths.}
\label{fig:data_analysis}
\end{figure}

\section{Framework}

Our evaluation framework focuses on assessing the effectiveness of different prompting strategies and program-based approaches for deterministic computation tasks. Unlike traditional model-centric designs, our framework emphasizes the interaction between natural language instructions, model outputs, and (when applicable) external execution.

\subsection{Task Pipeline}

All methods follow a unified pipeline:

\begin{center}
Natural Language Input $\rightarrow$ LLM Generation $\rightarrow$ (Optional Code Execution) $\rightarrow$ Final Answer
\end{center}

For standard prompting methods, the model directly generates the final answer in text form. For program-based approaches, the model first generates executable code, which is then executed to produce the final result.

\subsection{Prompt Design}

We use template-based prompts to support multiple task types under a unified interface. Each prompt consists of four key components:

\begin{itemize}
\item \texttt{task\_type}: Specifies the problem category (binary counting, longest substring, or arithmetic).
\item \texttt{instruction}: A natural language description sampled from a diverse instruction set.
\item \texttt{input\_data}: The sequence or expression to be processed.
\item \texttt{target\_char}: An optional field used for substring tasks.
\end{itemize}

All prompts explicitly constrain the output format to ensure consistent evaluation. For example, binary counting requires the format \texttt{0:x | 1:y}, while other tasks require a single integer output.

Different prompting strategies are implemented by modifying the reasoning instructions within the same template. Full prompt templates are provided in the Appendix.

\subsection{Prompting Approaches}

We evaluate five prompting strategies:

\begin{itemize}
\item \textbf{Plain Prompt}: The model directly produces the final answer given the instruction, without explicit reasoning guidance.

\item \textbf{Chain-of-Thought (CoT)}: The model is encouraged to generate intermediate reasoning steps before producing the final answer.

\item \textbf{Least-to-Most}: The task is decomposed into smaller subproblems, which are solved sequentially.

\item \textbf{Program-of-Thought (PoT)}: The model generates executable Python code instead of a direct answer, enabling external computation.

\item \textbf{Self-Consistency (SC)}: Multiple outputs are generated using stochastic decoding, and the final answer is selected via majority voting.
\end{itemize}

\subsection{Execution Mechanism}

For PoT, generated code is executed using a Python runtime. To ensure safe and reliable execution, we apply several constraints:

\begin{itemize}
\item Disallow unsafe operations such as \texttt{input()}, \texttt{sys.stdin}, and file I/O.
\item Execute code in a restricted environment with predefined variables (e.g., \texttt{input\_data}, \texttt{target\_char}).
\item Capture standard output as the final result.
\end{itemize}

This design allows the model to offload deterministic computation to an external interpreter, reducing reliance on internal reasoning.

\subsection{Self-Consistency Setup}

For Self-Consistency, we generate multiple outputs (5 samples per input) using temperature-based sampling ($T = 0.7$). Each output is parsed into a structured answer, and the final prediction is selected using majority voting.

\subsection{CodeT5-Based Approach}

In addition to prompting-based methods, we train a small code generation model, CodeT5-small, under a sequence-to-sequence framework.

The model takes as input the task description and generates a Python program that solves the task. The generated code is then executed to produce the final answer.

This approach can be viewed as a learned version of PoT, where the mapping from task description to executable program is optimized through supervised training rather than prompting.

\section{Experiment}

\subsection{Evaluation Settings}

We evaluate all methods under two settings:

\textbf{(1) Single-task evaluation.}  
We first evaluate on the binary counting task to establish baseline performance on a simple deterministic problem. We use 100 samples for standard prompting methods and 50 samples for Self-Consistency (SC) due to its higher computational cost. For SC, each sample is evaluated with 5 independent queries.

\textbf{(2) Mixed-task evaluation.}  
We then evaluate generalization across tasks by constructing a mixed dataset containing binary counting, longest substring, and arithmetic computation tasks. We sample 150 examples in total, evenly distributed across the three tasks (50 per task). For SC, we reduce the evaluation set to 90 samples (30 per task), with 5 queries per sample, resulting in 450 total API calls.

\subsection{LLM Evaluation}

We use the \texttt{gpt-oss-120b} model for all prompting-based evaluations under a zero-shot setting. We compare five prompting strategies:

\begin{itemize}
\item Plain Prompt
\item Chain-of-Thought (CoT)
\item Least-to-Most
\item Program-of-Thought (PoT)
\item Self-Consistency (SC)
\end{itemize}

For SC, we set the temperature to 0.7 to encourage output diversity and apply majority voting over 5 generated responses. 

For PoT, the model generates executable Python code, which is executed in a controlled environment. To ensure safe execution, we block potentially unsafe operations (e.g., \texttt{sys.stdin}, \texttt{input()}, file I/O).

\subsection{CodeT5 Training}

We additionally train a small code generation model, CodeT5-small, to compare against prompting-based approaches.

\textbf{Single-task training.}  
We train CodeT5 on the binary counting dataset (1000 samples). The dataset is split into 90\% training and 10\% validation.

\textbf{Mixed-task training.}  
We further train CodeT5 on the combined dataset of three tasks (3000 samples total). The same 90/10 split is used.

\textbf{Model configuration.}  
We use the pre-trained \texttt{Salesforce/codet5-small} model. The input consists of the task instruction and input sequence, while the output is the corresponding Python program.

\textbf{Training hyperparameters.}
\begin{itemize}
\item Learning rate: $5 \times 10^{-5}$
\item Batch size: 8 (train and evaluation)
\item Number of epochs: 6
\item Weight decay: 0.01
\item Evaluation strategy: per epoch
\end{itemize}

Sequence lengths are determined based on dataset analysis:
\begin{itemize}
\item Single-task: MAX\_INPUT\_LEN = 96, MAX\_TARGET\_LEN = 32
\item Mixed-task: MAX\_INPUT\_LEN = 160, MAX\_TARGET\_LEN = 83
\end{itemize}

Training is performed on a single Google Colab T4 GPU, with total training time approximately 3 minutes.

\subsection{Evaluation Metric}

We use exact-match accuracy as the evaluation metric. A prediction is considered correct only if it exactly matches the ground-truth answer. For arithmetic tasks, outputs such as \texttt{5} and \texttt{5.0} are treated as equivalent during evaluation.

\section{Results}

\subsection{Single-task Results (Binary Counting)}

\begin{center}
\begin{tabular}{l c}
\hline
Method & Accuracy \\
\hline
Plain Prompt & 0.56 \\
CoT & 0.57 \\
Least-to-Most & 0.40 \\
PoT & 1.00 \\
Self-Consistency & 0.64 \\
CodeT5 & 1.00 \\
\hline
\end{tabular}
\end{center}

In the single-task setting, standard prompting methods (Plain, CoT) achieve moderate accuracy (around 0.56–0.57), indicating that LLMs struggle with exact counting over long sequences. CoT provides minimal improvement, suggesting that explicit reasoning prompts do not enhance deterministic computation. Least-to-Most performs significantly worse due to error accumulation across steps.

Self-Consistency improves performance (0.64) via majority voting but remains unreliable and computationally expensive. In contrast, PoT and CodeT5 both achieve perfect accuracy by relying on executable computation.

\subsection{Mixed-task Results (Per-task Analysis)}

We report results separately for each task to highlight differences in task difficulty.

\subsubsection{Binary Counting}

\begin{center}
\begin{tabular}{l c}
\hline
Method & Accuracy \\
\hline
Plain Prompt & 0.54 \\
CoT & 0.52 \\
Least-to-Most & 0.48 \\
PoT & 1.00 \\
Self-Consistency & 0.73 \\
CodeT5 & 1.00 \\
\hline
\end{tabular}
\end{center}

Binary counting remains challenging for LLMs, requiring precise token-level reasoning over long sequences. Performance trends are consistent with the single-task setting.

\subsubsection{Longest Substring}

\begin{center}
\begin{tabular}{l c}
\hline
Method & Accuracy \\
\hline
Plain Prompt & 0.42 \\
CoT & 0.42 \\
Least-to-Most & 0.38 \\
PoT & 1.00 \\
Self-Consistency & 0.47 \\
CodeT5 & 1.00 \\
\hline
\end{tabular}
\end{center}

The longest substring task is the most difficult among the three. It requires maintaining continuous state across the entire sequence, making it more sensitive to local errors. All prompting-based methods perform poorly, and even Self-Consistency provides limited improvement.

\subsubsection{Arithmetic Computation}

\begin{center}
\begin{tabular}{l c}
\hline
Method & Accuracy \\
\hline
All Methods & 1.00 \\
\hline
\end{tabular}
\end{center}

All methods achieve perfect accuracy on arithmetic expressions. These problems are short, structured, and well-aligned with patterns seen during pretraining, making them significantly easier than sequence-based tasks.

\subsection{Overall Mixed Performance}

For completeness, we also report overall mixed-task accuracy:

\begin{center}
\begin{tabular}{l c}
\hline
Method & Accuracy \\
\hline
Plain Prompt & 0.65 \\
CoT & 0.65 \\
Least-to-Most & 0.62 \\
PoT & 1.00 \\
Self-Consistency & 0.73 \\
CodeT5 & 1.00 \\
\hline
\end{tabular}
\end{center}

However, this aggregate metric is dominated by the arithmetic task, which is trivially solved by all methods. As a result, mixed accuracy may overestimate the true capability of LLMs on harder deterministic problems.

\subsection{Analysis and Discussion}

The results reveal a clear hierarchy of task difficulty:

\begin{center}
Arithmetic $\ll$ Binary Counting $<$ Longest Substring
\end{center}

Arithmetic tasks in our benchmark are easily solved by LLMs, while sequence-based tasks requiring exact counting and state tracking remain challenging.

CoT does not meaningfully improve performance, suggesting that LLMs generate plausible reasoning traces without performing true computation. Least-to-Most further degrades performance due to error accumulation across intermediate steps.

Self-Consistency improves accuracy through repeated sampling but remains inefficient and unreliable, requiring multiple model calls per query.

In contrast, PoT achieves perfect performance by separating reasoning from computation and delegating execution to an external interpreter. Similarly, CodeT5 achieves perfect accuracy on the synthetic test set across all tasks after supervised training, demonstrating that deterministic computation can be reliably learned through code generation.

These findings suggest that external tools or specialized models are necessary for reliable exact computation, as LLMs alone are not sufficient for such tasks.

\section{Conclusion}

In this work, we investigated the effectiveness of prompting strategies for deterministic computation tasks. Our results show that LLMs struggle to produce exact and reliable outputs when relying solely on natural language reasoning. While techniques such as Chain-of-Thought and Least-to-Most prompting can generate plausible reasoning processes, they provide limited improvement in accuracy and remain prone to error accumulation. These results suggest that LLMs may generate plausible reasoning patterns without reliably performing exact symbolic computation.

Program-of-Thought offers a fundamentally different approach by separating reasoning from computation. By generating executable code and delegating execution to an external interpreter, PoT achieves perfect performance across tasks. Self-Consistency further improves robustness through repeated sampling and majority voting, but introduces substantial computational overhead without guaranteeing reliability.

We also demonstrate that a small domain-specific model, CodeT5-small, can achieve perfect accuracy on the synthetic test set across all tasks after lightweight training. This highlights that for deterministic problems, specialized models trained to generate executable programs can be both more accurate and more efficient than large general-purpose LLMs.

We note that our dataset is synthetic and focuses on well-defined deterministic tasks. Extending these findings to real-world noisy or long-horizon reasoning problems remains an important direction for future work.

Overall, our findings suggest that LLMs are not well-suited for exact computation in a purely text-based setting. Instead, combining LLMs with external tools or adopting task-specific models provides a more reliable and scalable solution. We hope this work motivates further research into integrating symbolic computation and program execution with language models.

\appendix
\renewcommand{\thesection}{\Alph{section}}
\section{Prompt Templates}

In this section, we provide the prompt templates used in our experiments. All prompts share a common structure with task-specific instructions and output constraints.

\subsection{Plain Prompt}

\begin{lstlisting}
You are a precise computation assistant.

Task type:
{task_type}

Instruction:
{instruction}

Input:
{input_data}

Target character if needed:
{target_char}

Rules:
- Do not estimate.
- Follow the instruction exactly.
- Return only the final answer.

Output format:
- If task_type is binary_count, return: 0:<count_0> | 1:<count_1>
- If task_type is longest_substring, return only the integer.
- If task_type is arithmetic, return only the integer.
\end{lstlisting}

\subsection{Chain-of-Thought (CoT)}

\begin{lstlisting}
You are a careful computation assistant.

Task type:
{task_type}

Instruction:
{instruction}

Input:
{input_data}

Target character if needed:
{target_char}

Think step by step to solve the task carefully.
Do not estimate.
After reasoning, return only the final answer.

Output format:
- If task_type is binary_count, return: 0:<count_0> | 1:<count_1>
- If task_type is longest_substring, return only the integer.
- If task_type is arithmetic, return only the integer.
\end{lstlisting}

\subsection{Least-to-Most Prompt}

\begin{lstlisting}
You are a careful computation assistant.

Task type:
{task_type}

Instruction:
{instruction}

Input:
{input_data}

Target character if needed:
{target_char}

Solve the task by decomposing it into smaller steps.

For binary_count:
- Split the sequence into smaller chunks.
- Count each chunk.
- Sum the counts.

For longest_substring:
- Scan the sequence from left to right.
- Track current consecutive length.
- Track maximum consecutive length.

For arithmetic:
- Apply standard operator precedence.
- Solve multiplication/division before addition/subtraction.

Return only the final answer.

Output format:
- If task_type is binary_count, return: 0:<count_0> | 1:<count_1>
- If task_type is longest_substring, return only the integer.
- If task_type is arithmetic, return only the integer.
\end{lstlisting}

\subsection{Program-of-Thought (PoT)}

\begin{lstlisting}
You are a precise Python programmer.

Task type:
{task_type}

Instruction:
{instruction}

Input:
{input_data}

Target character if needed:
{target_char}

Important:
- Do NOT use input(), sys.stdin, or file reading.
- The input is already available in the variable input_data.
- The target character is already available in the variable target_char.

Write Python code to solve the task.

Requirements:
- Use input_data as the input variable.
- If task_type is longest_substring, use target_char as the target character.
- Print only the final answer.
- Do not include explanations.
- Do not include markdown.
- Output only valid Python code.

Expected print format:
- binary_count: print(f"0:{count_0} | 1:{count_1}")
- longest_substring: print(max_consecutive)
- arithmetic: print(result)
\end{lstlisting}

\subsection{Self-Consistency Prompt}

\begin{lstlisting}
You are a precise computation assistant.

Task type:
{task_type}

Instruction:
{instruction}

Input:
{input_data}

Target character if needed:
{target_char}

Rules:
- Do not estimate.
- Solve carefully.
- Double-check the result.
- Return only the final answer.

Output format:
- If task_type is binary_count, return: 0:<count_0> | 1:<count_1>
- If task_type is longest_substring, return only the integer.
- If task_type is arithmetic, return only the integer.
\end{lstlisting}

\end{document}